\documentclass[pdflatex,sn-aps]{sn-jnl}

\usepackage{graphicx}
\usepackage{amsmath}
\usepackage{amssymb}
\usepackage{siunitx}
\usepackage{booktabs}
\usepackage{multirow}
\usepackage{tabularx}
\usepackage{algorithm}
\usepackage{algpseudocode}

\begin{document}
	\title{Petri Net Description of Biological Neural Circuits for Fast Hardware Prototyping}

	\author*[1,2]{\fnm{Carlo}\sur{daCunha}}\email{carlo.dacunha@njit.edu}
	\author[3]{\fnm{Rodrigo}\sur{Pena}}\email{penar@fau.edu}
	\author[4]{\fnm{Marcos}\sur{Turqueti}}\email{mturqueti@lbl.gov}
	
	\affil*[1]{\orgdiv{Helen and John C. Hartmann Department of Electrical Engineering},
	            \orgname{New Jersey Institute of Technology},
	            \orgaddress{
	            	\street{323 Martin Luther King Jr. Blvd.},
		        \city{Newark},
	                \postcode{07102},
	                \state{NJ},
	                \country{USA}}}

	\affil[2]{\orgdiv{School of Applied Engineering and Technology},
	            \orgname{New Jersey Institute of Technology},
	            \orgaddress{
	            	\street{323 Martin Luther King Jr. Blvd.},
		        \city{Newark},
	                \postcode{07102},
	                \state{NJ},
	                \country{USA}}}
	                
	\affil[3]{\orgdiv{Department of Biological Sciences},
	            \orgname{Florida Atlantic University},
	            \orgaddress{
	            	\street{5353 Parkside Drive},
		        \city{Jupiter},
	                \postcode{33458},
	                \state{FL},
	                \country{USA}}}
	
	\affil[4]{\orgdiv{Electronic Systems Group},
	            \orgname{Lawrence Berkeley National Laboratory},
	            \orgaddress{
	            	\street{1 Cyclotron Road.},
		        \city{Berkeley},
	                \postcode{94720},
	                \state{CA},
	                \country{USA}}}


	\abstract{
	 	Current approaches to simulating biological neural circuits, whether on general-purpose hardware or dedicated neuromorphic platforms, remain constrained by fixed-timestep numerical integration, hardware-imposed precision limits, and an inability to guarantee timing correctness for event-driven spiking dynamics under real-time constraints. Here, we propose a Petri net description of biological neural circuits that overcomes these limitations by modeling neurons, synapses, and spike events as a T-timed Petri net with formally verifiable timing semantics, enabling deadline-guaranteed real-time execution and analytically tractable correspondence to continuous-time leak-integrate-and-fire dynamics, independent of the underlying integration timestep. To test the model, we present the results of three simulated microcircuits: feedback inhibition, lateral inhibition, and hierarchical feature detector. The Petri neuron reproduces the expected dynamical signatures of each circuit while providing formally bounded timing guarantees throughout, with worst-case response times matching analytical predictions across all three cases.}

	\maketitle
	
	\section{Introduction}
	\label{sec:intro}
	Spiking neural networks (SNNs) are the third generation of artificial neural network models~\cite{maass1997networks}, distinguished from their rate-coded predecessors by their use of discrete spike events to encode and transmit information in the time domain. This event-driven computation paradigm endows SNNs with a natural affinity for temporal pattern recognition, reactive control, and energy-efficient inference on resource-constrained hardware~\cite{pfeiffer2018deep}. Neuromorphic platforms such as Intel Loihi\cite{davies2018loihi} and IBM TrueNorth~\cite{merolla2014million} exploit these properties at scale, and recent work has demonstrated real-time cortical simulation of 77,000 neurons on SpiNNaker at biological timescale~\cite{rhodes2019realtime}.
	
	The central challenge of deploying SNNs on general-purpose embedded hardware is that spike timing is not merely a performance metric but a computational variable. In biological neural circuits such as central pattern generators (CPGs)~\cite{marder2001central}, the precise timing of each spike determines motor output, sensory gating, and rhythmic coordination. A locomotor CPG that fires 10~ms late does not merely produce a slower gait but produces the wrong gait. This places SNN execution in the domain of hard real-time systems, where timing correctness is a functional requirement rather than a quality-of-service preference. Yet existing SNN simulators and inference engines, including Brian~2~\cite{stimberg2019brian}, SNNTorch~\cite{eshraghian2023training}, and PyNN~\cite{davison2008pynn}, are designed for scientific simulation rather than real-time deployment and provide no mechanisms for bounding worst-case spike delivery latency. Hardware schedulers for neuromorphic SoCs~\cite{balaji2019mapping} address throughput and energy efficiency but operate on synchronous-timestep models incompatible with continuous-time biological dynamics. Prior work on Petri net models of neural computation, including Spiking Neural P systems~\cite{ionescu2006spiking} and their Petri net translations~\cite{wang2010modeling}, has focused on computational expressiveness and Turing completeness rather than timing guarantees. The question of whether a given SNN configuration can be executed on a given hardware platform with bounded spike-delivery latency has not previously received a formal answer.
	
	To address this gap, we leverage formal real-time scheduling theory~\cite{liu1973scheduling} and timed Petri nets~\cite{ramchandani1974analysis}, which together provide the analytical tools needed to specify concurrent systems with timing constraints and derive schedulability conditions in terms of worst-case execution times and minimum inter-arrival periods. We introduce the \emph{Petri neuron}, a formal model of the leaky integrate-and-fire (LIF) neuron cast as a T-timed Petri net, and develop a complete schedulability analysis yielding a closed-form worst-case response time (WCRT) computable at design time from neuron parameters and platform-specific jitter bounds. Parameter mapping rules derived via Pad\'e approximation allow a designer to instantiate a Petri neuron directly from standard biophysical parameters.
	
	We validate the model on three biological microcircuits: feedback inhibition, lateral inhibition, and a hierarchical feature detector, implemented in C++ and cross-validated on an AMD x86 Linux workstation and a Raspberry Pi RP2040 microcontroller. The Petri neuron reproduces gamma-band rhythms in the feedback inhibition network. It provides a formal schedulability condition guaranteeing that inhibitory feedback arrives within a bounded number of cycles, preventing runaway excitation. The lateral inhibition circuit resolves winner-take-all competition with provably bounded latency, independent of input contrast. The hierarchical feature detector reproduces orientation selectivity consistent with the HMAX model and empirically confirms WCRT predictions. These results demonstrate that Petri net schedulability analysis translates directly into formal timing guarantees for neural circuits. These guarantees are neither available in continuous LIF models nor in existing SNN simulation frameworks.
	
	The remainder of this paper is organized as follows. Section~\ref{sec:model} defines the Petri neuron, establishes its structural and behavioral properties through invariant and reachability analysis, derives the LIF correspondence and WCRT bounds, characterizes platform jitter, and describes the event-driven implementation. Section~\ref{sec:results} presents the three microcircuit case studies. Section~\ref{sec:conclusions} concludes the paper and gives research directions.

	\section{Model}
	\label{sec:model}
	
	\subsection{Petri Neuron Definition}
	
\begin{figure}[htbp]
		\centerline{
			\includegraphics{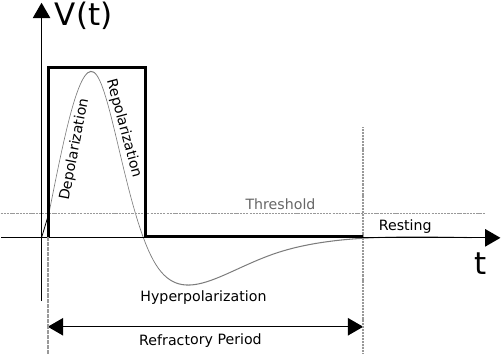}
		}
		\caption{Phases of a biological action potential. Depolarization (rapid voltage rise past threshold), repolarization (voltage returns toward rest), hyperpolarization (transient overshoot below resting potential), and the subsequent return to rest via ionic pump activity. The refractory period spans repolarization and hyperpolarization, during which the neuron is unresponsive (absolute) or requires elevated excitation to fire (relative). This cycle motivates the input-gating, threshold-firing, and recovery transitions of the Petri neuron in Fig.~\ref{fig:PetriNeuron}.}
		\label{fig:action_potential}
	\end{figure}

	A biological neuron in a rest state maintains a negative potential relative to its surroundings (roughly $-70$ mV) due to an ionic sodium and potassium imbalance and the membrane properties \cite{hodgkin1952quantitative,hodgkin1952propagation}. Upon receiving electrical stimulation, the membrane voltage depolarizes and reaches the critical threshold (around $-55$ mV). The neuron then enters a state in which ionic currents' dynamics cause a rapid voltage increase and further depolarization, leading to a stereotypical discharge called an action potential or spike. After reaching its peak, sodium channels close and potassium channels open, triggering a repolarization phase in which the membrane potential returns to its resting value, often a hyperpolarized state, which is typically overshooted, driving the voltage to more negative values ($\sim -75$ mV). During these phases, the biological neuron enters a two-stage refractory period in which it does not respond to external excitation: the relative refractory period and the absolute refractory period.  The absolute refractory period ends with the end of the hyperpolarization, but it takes a higher excitation to fire the neuron during this relative refractory period. Finally, sodium-potassium pumps return the neuron to its resting state. This voltage spike then propagates through the axon until reaching its terminals to communicate with another neuron. This process is illustrated in Fig. \ref{fig:action_potential}.
		
\begin{figure}
		\centering
		\includegraphics[scale=1.25]{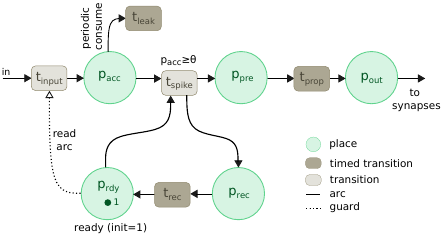}
		\caption{Petri net representation of the Petri neuron, mapping accumulation, threshold firing, axonal propagation, and refractory recovery onto places and transitions.}
		\label{fig:PetriNeuron}
	\end{figure}

	To model the behavior of a biological spiking neuron, we define a Petri neuron as a Petri Net, shown in Fig. \ref{fig:PetriNeuron}. The transition $t_{input}$ only allows the Petri neuron to receive a token (mimicking an excitation pulse) if it is not in the refractory state. Once the token is received, it accumulates in the place $p_{acc}$. However, in the biological neuron, charge leaks through the membrane, and this is modeled by the transition $t_{leak}$ that periodically consumes tokens. Once the number of tokens reaches a threshold $\theta$, they are instantly consumed by the transition $t_{spike}$. This transition also consumes a token from the place $p_{rdy}$ if it has one. $p_{rdy}$ is a place that indicates that the neuron is not in its refractory state. The transition $t_{spike}$ creates a token at the place $p_{pre}$, indicating that the neuron is ready to propagate a spike through its axon. It also creates a token at location $p_{rec}$, indicating that the neuron is recovering from its refractory state. Once the refractory state is over, the timed transition $t_{rec}$ consumes the token from $p_{rec}$ and creates a new token in $p_{rdy}$. Parallel to this recovery process, the timed transition $t_{prop}$ will consume the token from $p_{pre}$ after propagation is completed and create one at $p_{out}$, which then connects to other neurons.
	
	\subsection{Structural Analysis}
	
	The structural and behavioral properties of the Petri neuron are established by analyzing its incidence matrix, invariants, reachability set, and classical Petri net properties. 
	
	The incidence matrix $\mathbf{C}=\mathbf{C}^+-\mathbf{C}^-$, where $\mathbf{C}^+$ is the post-incidence (tokens produced) and $\mathbf{C}^-$ is the pre-incidence (tokens consumed). Read arcs do not appear in the incidence matrix since they neither produce nor consume tokens. For the Petri neuron, the incidence matrix is:
	
	\begin{table}[h]
	\centering
	\renewcommand{\arraystretch}{1.4}
	\begin{tabular}{l|ccccc}
	\toprule
	 & $T_{input}$ & $T_{spike}$ & $T_{prop}$ & $T_{rec}$ & $T_{leak}$ \\
	\midrule
	$p_{acc}$ & $1/0/+1$  & $0/\theta/-\theta$   & $0/0/0$  & $0/0/0$  & $0/1/-1$ \\
	$p_{rdy}$ & $0/0/0$   & $0/1/-1$   & $0/0/0$  & $1/0/+1$ & $0/0/0$  \\
	$p_{pre}$ & $0/0/0$   & $1/0/+1$  & $0/1/-1$ & $0/0/0$  & $0/0/0$  \\
	$p_{out}$ & $0/0/0$   & $0/0/0$    & $1/0/+1$ & $0/0/0$  & $0/0/0$  \\
	$p_{rec}$ & $0/0/0$   & $1/0/+1$  & $0/0/0$   & $0/1/-1$ & $0/0/0$  \\
	\bottomrule
	\end{tabular}
	\caption{Incidence matrix with entries $C^+/C^-/C$. $\theta$ denotes the current token count of $p_{acc}$ at firing time (flush arc).}
	\label{tab:incidence}
	\end{table}
	
	From the incidence matrix, we computed the place-coupling Laplacian matrix (PCLM) $\mathbf{C}\mathbf{C}^\top$ to find functional clustering properties of the Petri neuron and the transition-coupling Laplacian matrix (TCLM) $\mathbf{C}^\top\mathbf{C}$ to find concurrency groups.
	
	The geometric multiplicity of the zero eigenvalue of $\mathbf{C}\mathbf{C}^\top$ is $1$, indicating the presence of a single invariant. The eigenvector of the PCLM corresponding to the null eigenvalue is $\begin{bmatrix}0 & -0.71 & 0 & 0 & -0.71\end{bmatrix}$, which is proportional to $\begin{bmatrix}0 & 1       & 0 & 0  & 1\end{bmatrix}$. According to the p-invariant therem ($\mathbf{x}\cdot\mathbf{M}=\mathbf{x}\cdot\mathbf{M}_0=M_0(p_{rdy})=1$) yields the mutual exclusion invariant $M(p_{rdy})+M(p_{rec})=1$. This formally proves that the Petri neuron is always either ready or recovering.
	
	The eigenvector of the TCLM corresponding to the null eigenvalue is $\begin{bmatrix}0.71 & 0 & 0 & 0 & 0.71\end{bmatrix}$, which implies that the only possibility for the network to return to a previous configuration is with $t_{input}$ and $t_{leak}$ firing the same number of times, which is biologically plausible. However, if the neuron starts with a combined total of $N$ tokens in $p_{rdy}$ and $p_{rec}$, any marking that requires more or fewer than $N$ tokens in those places is mathematically unreachable.
	
	The Fiedler value $\lambda_2$ of the symmetric, positive semidefinite matrix $\mathbf{C}\mathbf{C}^\top$ is relatively small. While the Fiedler value traditionally characterizes the algebraic connectivity and diffusion speed of undirected graphs, we extend this interpretation to the Petri Net setting to quantify structural bottlenecks. In this context, a small $\lambda_2$ indicates that the network synchronizes slowly, meaning that a single-transition bottleneck constrains the flow of tokens and the transition firings between the two major sub-components. The corresponding structural eigenvector $\mathbf{x}=\begin{bmatrix}-0.13 & -0.13 & -0.62 & -0.76 & 0.13\end{bmatrix}$ confirms this structural divide: $p_{pre}$ and $p_{out}$ serve as the dominant places of one cluster module. In contrast, $p_{rec}$ sits entirely on the opposing side of the partition. Consequently, the worst-case structural latency of the system is dominated by the token-clearing time of the $p_{pre}/p_{out}$ module before it can successfully synchronize with the rest of the network. 
	
	The equivalent TCLM eigenvector is $\begin{bmatrix}-0.31 & -0.57 & -0.31 & -0.62 & 0.31\end{bmatrix}$. This eigenvector isolates $t_{leak}$ in its own partition ($x_{leak}=0.31$), while $t_{spike}$ ($x_{spike}=-0.57$) and $t_{rec}$ ($x_{rec}=-0.62$) are heavily involved in a second module. Because the eigenvector $\mathbf{x}$ lies in the kernel of the incidence matrix $\mathbf{M}$, the subnets governing the evolution of $t_{spike}$ and $t_{rec}$ are topologically isomorphic. Consequently, under an event-driven, token-passing execution model, the firing rates and token conservation laws for both partitions evolve under identical state-transition systems.
	Consequently, these symmetric components can be mapped to separate threads carrying identical structural weight, exerting identical load profiles on the resource pool.
	
	Two eigenvalues are equal ($\lambda_3=\lambda_4=2$), which implies structural symmetry of the Petri neuron. The rows corresponding to $t_{input}$ and $t_{leak}$ in the third and fourth TCLM eigenvectors are perfectly antisymmetric, confirming the t-invariance. Moreover, the row corresponding to $t_{spike}$ is zero, placing it at the behavioral center of the network; algebraically, this zero-component indicates that $t_{spike}$ acts as the invariant topological pivot or cut-vertex between the symmetric subnet partitions. This central role is confirmed by the TCLM eigenvector corresponding to the spectral radius ($\lambda_5\approx 5.83$), $\mathbf{x}_5=\begin{bmatrix}0.21 & -0.82 & 0.21 & 0.43 & -0.21\end{bmatrix}$. Because $\mathbf{x}_5$ is the Perron-Frobenius vector, its components correspond to the structural network centrality, thereby proving that $t_{spike}$ (exhibiting the highest absolute magnitude of $0.82$) is the dominant driver of token throughput.
	The corresponding PCLM eigenvector, $\begin{bmatrix}0.52 & 0.52 & -0.43 & 0.09 & -0.52\end{bmatrix}$, indicates that $p_{acc}$, $p_{rdy}$, and $p_{rec}$ form the core infrastructure supporting this central engine.
	
	\subsection{Reachability and Liveness}
	
	$p_{pre}$ and $p_{out}$ are downstream, forward-moving places that do not feed tokens back into the control loop and are, consequently, structurally unbounded. To isolate them during a reachability analysis, we apply a projection matrix $P\in\{0,1\}^{3\times 5}$ that restricts attention to the bounded core places $\{p_{acc},p_{rdy},p_{rec}\}$. This left-linear transformation yields a structural monomorphism mapping the original net to a bounded core in which $p_{acc}$ is $\theta$-bounded and $p_{rdy}$ and $p_{rec}$ are 1-bounded.
	
	By solving the state equation $M=C\sigma+M_0$ under the projection $M_{core}=PM$, subject to the constraints $\sigma\geq 0$, $M\geq 0$, and $\mathbf{y}_{core}\cdot M_{core}>0$, we find the valid vectors that define the core reachability set. These correspond to three fundamental operational states: a resting state ($\sigma_1=\begin{bmatrix}0 & 1 & 0\end{bmatrix}$), an integrating state ($\sigma_2=\begin{bmatrix}\theta & 1 & 0\end{bmatrix}$), and a refractory recovering state ($\sigma_3=\begin{bmatrix}0 & 0 & 1\end{bmatrix}$).
	
	The state transitions follow a cyclic topology characterized by the sequence: $\xrightarrow{t_{spike}}$ integrating $\xrightarrow{t_{rec}}$ resting. Because every state in this sequence is reachable from any other via valid transition firings, the core network contains no deadlocks, and the Petri net is structurally live.
	
	\subsection{Correspondence with the Continuous LIF Model}
	\label{sec:lif_correspondence}
	
	The Petri neuron is a discrete-token abstraction of the leaky
	integrate-and-fire (LIF) model. The continuous membrane potential
	$V(t)$ is replaced by the token count $M(p_{acc})$, and the guard condition replaces the
	continuous threshold crossing
	$M(p_{acc}) \geq \theta$. This section establishes the relationship
	between the two models, identifies where they agree, and states
	explicitly where they differ.
		
	The subthreshold dynamics of the LIF neuron are governed by:
	
	\begin{equation}
		\tau_mdV(t)=-(V-RI)dt,
	\end{equation}
	where $V(t)$ is the membrane potential, $R$ is the membrane resistance, $I$ is the input current, and $\tau_m=RC$ is the membrane time constant. Assuming that the membrane voltage is continuously differentiable, its solution with $V(0)=0$ is:
	
	\begin{equation}
		V(t)=RI\left(1-e^{-t/\tau_m}\right).
	\end{equation}
	Once the membrane potential reaches a threshold voltage ($V_t$), the neuron fires and enters a refractory period. Rewriting this equation for frequency, one finds:
	
	\begin{equation}
		f_{LIF}=\left[\tau_{ref}-\tau_m\ln\left(1-\frac{V_{th}}{RI}\right)\right]^{-1}
		\label{eq:firing_rate_LIF}
	\end{equation}
	if $I>V_t/R$ and $0$ otherwise.
	
	In the Petri neuron, the accumulator grows in discrete steps of size $w$ every $T$ seconds, reduced by one token every $T_{leak}$ seconds. The resulting marking is:
	
	\begin{equation}
		M(p_{acc})=\max\left(0,\left\lfloor\frac{t}{T}\right\rfloor\cdot w-\left\lfloor\frac{t}{T_{leak}}\right\rfloor\right),
		\label{eq:Mcond}
	\end{equation}
	where $T$ is the inter-pulse interval.
	
	Applying the same rationale, for a threshold $\theta$, the firing frequency of the Petri neuron is:
	
	\begin{equation}
		f_{Petri}=\left(\tau_{ref}+\left[\frac{\theta}{w/T-1/T_{leak}}\right]\right)^{-1}.
		\label{eq:fPetri}
	\end{equation}
	
	To match parameters of the Petri neuron with those of the LIF, we apply the Pad\'e approximation to the log function in Eq.~\ref{eq:firing_rate_LIF}:
	
	\begin{equation}
		f_{LIF}\approx\left(\tau_{ref}+\left[\frac{CV_{th}}{I-V_{th}/R}\right]\right)^{-1}.
		\label{eq:fLIF}
	\end{equation}
	
	Comparing Eqs. \ref{eq:fLIF} and \ref{eq:fPetri}, we see that $\theta$ behaves like an accumulated charge. Therefore, we normalize all charges by the constant $q=CV_{th}/\theta$ and obtain the following design rules for the Petri neuron: 1) Freely choose $\theta$, 2) Make $w=\lceil TI/q\rceil$, where $T$ behaves as a sampling frequency, and 3) Make $T_{leak}=Rq/V_{th}$.
	
\begin{figure}
		\centering
		\includegraphics{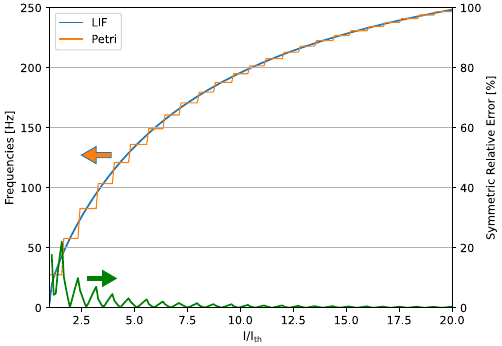}
		\caption{Firing rate as a function of input excitation $I/I_{th}$ for the LIF model (Eq.~\ref{eq:firing_rate_LIF}) and the Petri neuron (Eq.~\ref{eq:fPetri}), parameterized via the Pad\'e mapping. Agreement is close for spontaneous and moderate firing ($f<
        50$~Hz), with relative error shrinking further as $I$ increases into the high-frequency regime.}
		\label{fig:frequencies}
	\end{figure}

	For a typical cortical pyramidal neuron, $V_{th}\approx 20$ mV above the resting potential, $R\approx100$ M$\Omega$, and $C=100$ pF \cite{gerstner2002spiking,hodgkin1952quantitative}, which gives a membrane time constant $\tau_m\approx10$ ms. If we set $\theta=5$, for example, $q=8\times 10^{-13}$, $T_{leak}=4$ ms, and $w$ ranges from $2$ to $25$ for $I/I_{th}$ between $0$ and $20$. Both the frequencies of the LIF model and the Petri neuron are shown in Fig. \ref{fig:frequencies}. The Petri neuron shows symmetric relative errors less than $20$\% for spontaneous and active firing ($f<50$ Hz, $I<5I_{th}/2$) and this error asymptotically approaches zero for intense firing ($>100$ Hz, $I>5I_{th}/2$). The sensitivity of the firing rate of the Petri neuron with respect to changes in weight corroborates this:
			
	\begin{equation}
		\frac{\mathrm{d}f}{\mathrm{d}w} =
		\frac{T\,\theta}{\left[(w_{eff})\,\tau_{ref}
		+ T\,\theta\right]^2},
	\end{equation}
	which is approximately $(T\theta)^{-1}$ near rheobase and decreases quadratically with $\theta$. Consequently, neurons with smaller thresholds are more sensitive to weight modulation.
		
	From Eq. \ref{eq:Mcond}, a spiking event is triggered the moment the accumulation threshold is reached, satisfying $M(p_{acc})\geq\theta$. To derive the analytical bounds, we approximate the floor functions using the relation $\lfloor x\rfloor\approx x-\epsilon$, where the phase offset $\epsilon\in[0,1)$ accounts for the asynchronous alignment between the continuous timeline and the discrete arrivals of the inputs. This yields the continuous-time accumulation approximation:
	
	\begin{equation}
		t\approx\frac{M(p_{acc})+\epsilon_{in}\cdot w-\epsilon_{leak}}{\frac{w}{T}-\frac{1}{T_{leak}}}.
	\end{equation}
	The best-case response time ($t_{BCRT}$) occurs under perfect phase alignment, where an incoming spike lands exactly on the sampling clock edge ($\epsilon_{in}=0$), causing the accumulator to reach exactly $\theta$. Conversely, the worst-case response time ($t_{WCRT}$) occurs when an input spike arrives just after a clock edge ($\epsilon_{in}\rightarrow 1$). In this worst-case boundary condition, the token accumulation is delayed by an entire cycle, effectively forcing the net to accumulate an additional quantum of weight $w$ (setting the effective threshold boundary to $\theta+w$) before the threshold condition can register at the next epoch.
	
	Factoring out $T$ from the denominator yields the formal boundary equations:
	
	\begin{equation}
	  t_{BCRT} = \frac{\theta}{w-T/T_{leak}}\,T,
	  \qquad
	  t_{WCRT} = \frac{\theta + w}{w-T/T_{leak}}\,T,
	  \label{eq:bcrt_wcrt}
	\end{equation}
	which gives a temporal jitter bounded by:
	
	\begin{equation}
		\mathcal{J}_0=t_{WCRT}-t_{BCRT}=\frac{w}{w-T/T_{leak}}T.
	\end{equation}	
	For typical human cortical neurons with an injection current of $300$ pA, the jitter caused by channel and synaptic noise is typically less than $5$ ms. Therefore, the sampling period $T$ must be around $1$--$2$ ms for the Petri neuron.
	
	\subsection{Platform Characterization and Hardware Jitter}
	
	On top of the natural jitter, hardware jitter $\epsilon$ needs to be added, so that the total jitter is $\mathcal{J}=\mathcal{J}_0+\epsilon$. We characterize two platforms at opposite ends of the deployment spectrum: a general-purpose 3.8 MHz AMD Ryzen 9 3900X x86-64 workstation running Linux for general-purpose simulations, and a Raspberry Pi RP2040 microcontroller for deployment. While the latter has a highly deterministic microarchitecture with in-order execution, the former is profoundly non-deterministic in its execution timing and behavior due to out-of-order execution, speculative execution, branch prediction, dynamic scheduling, and other hardware optimizations. Moreover, the standard Linux kernel is not a deterministic operating system (OS).
				
	For each platform, we measured the signed timing error
	$e_k = t_{\mathrm{actual}}^{(k)} - d_j$ over $N = 10{,}000$ trials for
	five nominal delays $d_j \in \{1, 5, 10, 50, 100\}$~ms. The jitter bound
	is taken as the worst-case observed absolute error across all nominal delays
	and operating conditions:
	
	\begin{equation}
	  \epsilon_j = \max_{k,\, d_j} \left| t_{\mathrm{actual}}^{(k)} - d_j \right|
	  \label{eq:epsilon}
	\end{equation}
	
	On the AMD platform, we used a POSIX interval timer to deliver wakeups. In contrast, on the RP2040 platform, we used the direct hardware timer peripheral, which fires a CPU interrupt at the programmed target time with no OS involvement. The timer runs at exactly 1 MHz, giving a nominal resolution of 1 $\mu$s. Jitter was measured by reading the timer register as the first instruction of the interrupt service routine and comparing the result against the programmed target. Samples were streamed over USB to the host and analyzed offline.
	
	Table \ref{tab:jitter} shows all values for both platforms. RP2040 provides a consistent jitter of 1 $\mu$s, corresponding to quantization at the timer resolution floor. Moreover, the mean error of approximately 30 ns is consistent with the Cortex-M0+ interrupt latency of 6 cycles $= 48$ ns at 125 MHz. Conversely, the non-deterministic AMD-Linux platform exhibits a jitter that is delay-dependent, reaching a maximum value of $\sim905$ $\mu$s, which is comparable with the sampling period of the Petri neuron. Consequently, the AMD-Linux platform is only adequate to simulate short neurons with recovery times close to 1 ms.
	
	\begin{table}[t]
	  \centering
	  \caption{AMD Linux POSIX timer and RP2040 time jitter. All values in $\mu$s. $\epsilon$ denotes the worst-case absolute error ($N = 10{,}000$).}
	  \label{tab:jitter}
	  \small
	  \begin{tabularx}{\columnwidth}{l S[table-format=3.0] S[table-format=3.1] S[table-format=3.1] S[table-format=3.1] S[table-format=4.1]}
	      \toprule
	      \shortstack[l]{Machine}
	          & {\shortstack{Nom.\\(ms)}}
	          & {Mean}
	          & {Std}
	          & {p99}
	          & {$\epsilon$ (max)} \\
	      \midrule
	      \multirow{5}{*}{AMD}
	          &   1 &   9.7 &   2.3 &   16.9 &   95.9 \\
	          &   5 & 171.4 & 221.6 &  717.2 &  901.9 \\
	          &  10 & 207.4 & 227.7 &  728.2 &  917.7 \\
	          &  50 & 198.6 & 234.7 &  728.6 &  885.0 \\
	          & 100 & 252.0 & 240.8 &  730.0 &  905.3 \\
	      \midrule
	      \multirow{5}{*}{RP2040}
	          & 1 & 0.03 & 0.17 & 1.00 & 1.00 \\
	          & 5 & 0.03 & 0.16 & 1.00 & 1.00 \\
	          & 10 & 0.04 & 0.18 & 1.00 & 1.00 \\
	          & 50 & 0.02 & 0.14 & 1.00 & 1.00 \\
	          & 100 & 0.03 & 0.17 & 1.00 & 1.00 \\
	      \bottomrule
	  \end{tabularx}
	\end{table}
		
	\subsection{Network of Petri Neurons}
	A network of $N$ Petri neurons is defined by a set of neurons $\mathcal{N}=\{n_1,n_2,\hdots,n_N\}$ each parameterized by its threshold $\theta_i$, refractory delay $d_{rec}^{(i)}$, and propagation delay $d_{prop}^{(i)}$ and a set of weighted directed connections $\mathcal{C}\subset\mathcal{N}\times\mathcal{N}$, where each connection $(n_i,n_j)$ carries a synaptic weight $w_{ij}\in\mathbb{Z}^+$. 	The connection from $n_i$ to $n_j$ is realized by a synaptic transition $t_{syn}^{(ij)}$ such that $p_{out}^{(i)}\overset{t_{syn}^{(ij)}}{\rightarrow}p_{acc}^{(j)}$ carries weight $w_{ij}$, depositing $w_{ij}$ tokens into the downstream accumulation place per spike and consuming the token from $p_{out}^{(i)}$. Each connection also carries an associated synaptic delay $d_{syn}^{(ij)}$ representing the transmission time along the synapse.
	
	\subsection{Implementation}
	To implement the Petri neuron, we created a simple class consisting of procedures to add a spike, propagate a spike, recover the neuron from the refractory period, and periodically leak tokens. Moreover, we include a priority queue $Q$ that processes these neuronal events. The neuronal procedures are shown in Algorithm \ref{alg:PetriNeuron}, while the event-driven execution of the t-timed Petri network is shown in Algorithm \ref{alg:scheduler}.
	
	\begin{algorithm}[t]
	\caption{Transition firing rules for neuron $n$}
	\label{alg:PetriNeuron}
	\begin{algorithmic}[1]
	\Procedure{AddSpike}{$n, w, t$}
	    \If{$M_n(p_{\mathrm{rdy}}) > 0$} \Comment{accumulate}
	        \State $M_n(p_{\mathrm{acc}}) \gets M_n(p_{\mathrm{acc}}) + w$
	    \EndIf
	    \If{$M_n(p_{\mathrm{acc}}) \ge \theta_n$ \textbf{and} $M_n(p_{\mathrm{rdy}}) > 0$} \Comment{fire}
	        \State $M_n(p_{\mathrm{acc}}) \gets 0$;\; $M_n(p_{\mathrm{rdy}}) \gets 0$
	        \State $M_n(p_{\mathrm{rec}}) \gets 1$;\; $M_n(p_{\mathrm{pre}}) \gets 1$
	        \State \Call{Push}{$Q,\, (t + t_{\mathrm{prop}},\, \textsc{Propagation},\, n)$}
	        \State \Call{Push}{$Q,\, (t + t_{\mathrm{rec}},\, \textsc{Recovery},\, n)$}
	    \EndIf
	\EndProcedure
	\Statex
	\Procedure{Propagate}{$n, t$}
	    \If{$M_n(p_{\mathrm{pre}}) > 0$ \textbf{and} $t > \tau_{\mathrm{pre}} + t_{\mathrm{prop}}$}
	        \State $M_n(p_{\mathrm{pre}}) \gets 0$
	        \ForAll{$s \in \mathcal{S}_n$}
	            \State \Call{Push}{$Q,\, (t + s.\delta,\, \textsc{Spike},\, s.\mathrm{target},\, s.w)$}
	        \EndFor
	    \EndIf
	\EndProcedure
	\Statex
	\Procedure{Recover}{$n, t$}
	    \If{$M_n(p_{\mathrm{rec}}) > 0$ \textbf{and} $t > \tau_{\mathrm{rec}} + t_{\mathrm{rec}}$}
	        \State $M_n(p_{\mathrm{rec}}) \gets 0$;\; $M_n(p_{\mathrm{rdy}}) \gets 1$
	    \EndIf
	\EndProcedure
	\Statex
	\Procedure{Leak}{$n, t$}
	    \If{$M_n(p_{\mathrm{acc}}) > 0$ \textbf{and} $t > \tau_{\mathrm{leak}} + t_{\mathrm{leak}}$}
	        \State $M_n(p_{\mathrm{acc}}) \gets M_n(p_{\mathrm{acc}}) - 1$
	    \EndIf
	    \State \Call{Push}{$Q,\, (t + t_{\mathrm{leak}},\, \textsc{Leak},\, n)$} \Comment{periodic reschedule}
	\EndProcedure
	\end{algorithmic}
	\end{algorithm}
	
	\begin{algorithm}[t]
	\caption{Event-driven execution of the T-timed Petri net network}
	\label{alg:scheduler}
	\begin{algorithmic}[1]
	\Require Min-heap $Q$ of events $e = (\tau, \mathrm{kind}, n, w)$ keyed by timestamp $\tau$; neuron set $\mathcal{N}$; synapse lists $\{\mathcal{S}_n\}$
	\Procedure{ProcessEvents}{$t$}
  \While{$Q \neq \emptyset$ \textbf{and} $\min(Q).\tau \le t$}
      \State $e \gets \Call{Pop}{Q}$;\quad $n \gets e.\mathrm{id}$
      \If{$e.\mathrm{kind} = \textsc{Spike}$}
          \State \Call{AddSpike}{$n, e.w, t$}
      \ElsIf{$e.\mathrm{kind} = \textsc{Propagation}$}
          \State \Call{Propagate}{$n, t$}
      \ElsIf{$e.\mathrm{kind} = \textsc{Recovery}$}
          \State \Call{Recover}{$n, t$}
      \ElsIf{$e.\mathrm{kind} = \textsc{Leak}$}
          \State \Call{Leak}{$n, t$}
      \EndIf
  \EndWhile
	\EndProcedure
	\end{algorithmic}
	\end{algorithm}	
	
	\section{Results}
	\label{sec:results}
	
	\subsection{Feedback Inhibition}
	Feedback inhibition networks are the primary mechanism for self-regulation and gain control in the nervous system. It consists of excitatory cells (E), which are typically type A, L5 pyramidal neurons that use glutamate as the excitatory neurotransmitter, and inhibitory cells (I), which are typically Chrna2-expressing Martinotti (SST+) neurons that use gamma-aminobutyric acid (GABA) as the inhibitory neurotransmitter. 
	
\begin{figure}
		\centering
		\includegraphics{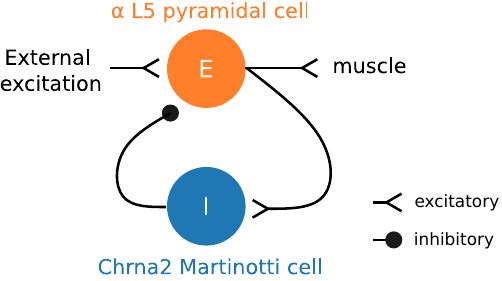}
		\caption{Feedback inhibition network. Chevron indicates an excitatory synapse, while the small black disks indicate inhibitory synapses.}
		\label{fig:PING}
	\end{figure}

	External input excites E neurons, causing them to fire. Their axons bifurcate with one branch exciting other circuits and a secondary collateral branch looping back locally to terminate on an I interneuron. The interneuron fires in response, sending inhibitory signals back onto the exact same E neurons that stimulated it, hyperpolarizing their membranes and suppressing their ability to keep firing. The feedback inhibition configuration is shown in Fig. \ref{fig:PING}
	
	Because of the built-in synaptic delay, the network naturally begins to pulse rhythmically. These loops, known as Pyramidal Interneuron Network Gamma (PING), are the direct source of the gamma rhythms ($30$--$80$ Hz) observed in electroencephalogram (EEG) readings during intense cognitive focus and memory processing.
	
	Table \ref{tab:finh} shows the parameters of E and I neurons in a typical feedback inhibition network.
	
	\begin{table}[t]
	    \centering
	    \caption{Neuronal parameters of excitatory (E) and inhibitory (I) neurons in a typical feedback-inhibition network.}
	    \small
	    \begin{tabularx}{\columnwidth}{l >{\centering\arraybackslash}X >{\centering\arraybackslash}X}
	        \toprule
	        Parameter & E & I \\
	        \midrule
	        $R$ [M$\Omega$]            & 30--60  & 337        \\
	        $C$ [$\times 10^{-10}$\,F] & 1.3--4  & 0.44--0.74 \\
	        $\tau$ [ms]                & 8--12   & 15--25     \\
	        $L$ [mm]                   & 1--3    & 4--8       \\
	        \bottomrule
	    \end{tabularx}
	    \label{tab:finh}
	\end{table}
	
\begin{figure}
		\centering
		\includegraphics{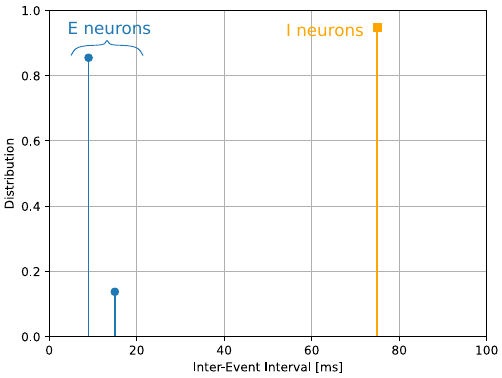}
		\caption{Inter-event interval histogram for the firing of E and I neurons in a feedback inhibition network simulated using Petri neuron.}
		\label{fig:EI}
	\end{figure}

	Neuron thresholds tend to be very similar. Therefore, we can arbitrarily set $\theta=5$ for both E and I neurons. From $T_L=\tau/\theta$, we set 2 ms for E, and 4 ms for I neurons. For unmyelinated cortical axons, the spike speed is between 0.3--0.5 m/s, which sets the propagation delay to 4 ms for E cells, and 17 ms for I cells. Finally, the inter spike interval (ISI), which relates to the refractory time, is between 1.5 ms and 2.0 ms for E neurons, and between 4.0 ms and 5.0 ms for I neurons. Figure \ref{fig:EI} shows the inter-event interval histogram (IEIH) obtained using the Petri neuron under this configuration and two synapses. One synapse is excitatory (weight = 3) connecting E neuron to I neuron through a delay of 1 ms. The other synapse is inhibitory (weight = $-5$) connecting I neuron back to E neuron through a delay of 3 ms. In the simulation, E neurons fire every 9 ms (111 Hz) 86 \% of the time and every 15 ms (67 Hz) 14 \% of the time, while all I neurons fire every 75 ms (13 Hz).
	
	The loop latency from an E-spike to the arrival of I feedback back onto E is $T_{loop}=d_{E\rightarrow I}+(t^I_{prop}+1)+d_{I\rightarrow E}=1+18+3=22$ ms, where the $+1$ term reflects the deterministic one-timestep penalty introduced by the strict inequality in the T-timed Petri net propagation guard. Since the steady-state E inter-spike interval is 9 ms, the first $k=\lceil T_{loop}/ISI_E\rceil=3$ spikes are uninhibited, constituting a bounded transient before feedback locks in. Once steady state is reached, the schedulability condition $T_{loop}<k$. $ISI_E^{steady}=27$ ms is satisfied, guaranteeing that inhibitory feedback arrives within a bounded number of cycles and preventing runaway excitation. This constitutes a formal real-time guarantee on the stability of the feedback inhibition loop, expressible directly in terms of the Petri net timing parameters.
	
	\subsection{Lateral (Afferent) Inhibition}
\begin{figure}
		\centering
		\includegraphics[scale=0.35]{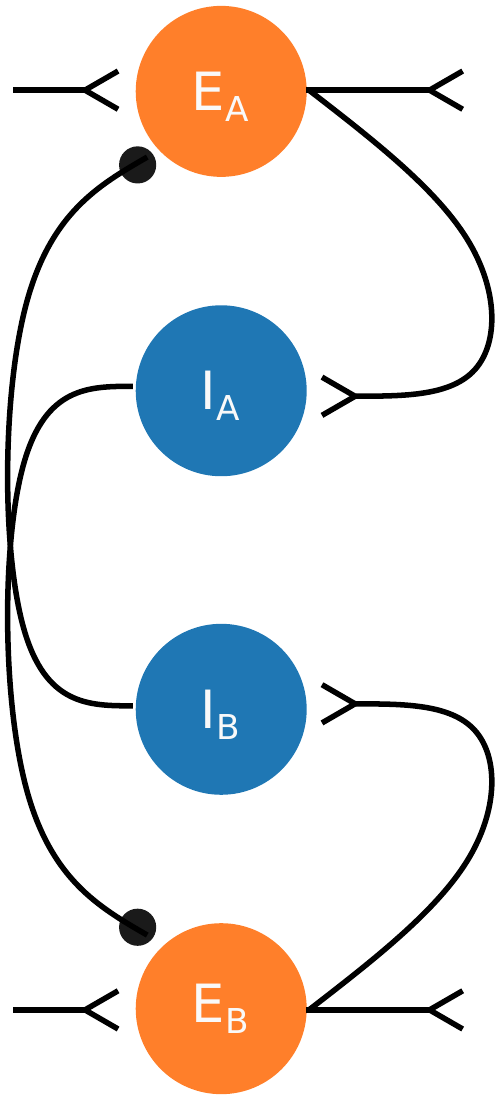}	
		\caption{Lateral inhibition microcircuit motif with excitatory neurons $E_{A,B}$ and inhibitory interneurons $I_{A,B}$.}
		\label{fig:lateral}
	\end{figure}

	Another important microcircuit motif is lateral inhibition, which is responsible for contrast enhancement and edge detection. This circuit is found universally in the retina as horizontal cells and in the olfactory bulb as mitral and granule cells. As shown in Fig. \ref{fig:lateral}, the lateral inhibition motif consists of two excitatory neurons that collateralize to inhibitory interneurons, which directly inhibit reciprocal excitatory neurons. This introduces competitive dynamics based purely on input contrast, with the excitatory neurons firing proportionally to the differential strength of the inputs.
	
\begin{figure}
		\centering
		\includegraphics{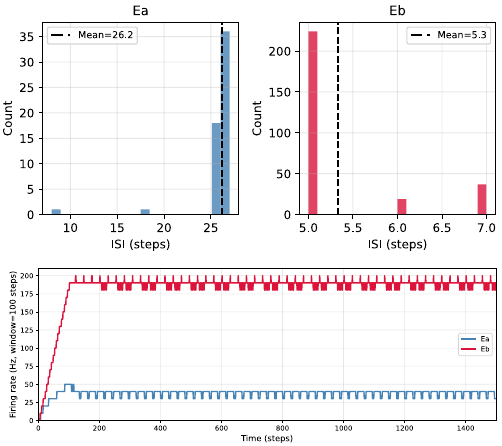}
		\caption{IEIH of excitatory neurons and their firing rates as a function of time.}
		\label{fig:rates}
	\end{figure}

	Figure \ref{fig:rates} shows the firing rates of excitatory neurons $E_{A, B}$ as well as their IEIH for the input of $E_B$ sending twice as many tokens per time as for the input of $E_A$.
	
	The competition resolution time, defined as the latency from stimulus onset to the arrival of inhibitory suppression at the losing channel, is bounded by $T_{comp}=d_{E\rightarrow I}+(t_{prop}^I+1)+d_{I\rightarrow E_{null}}=2+3+4=9$ ms, where the $+1$ term again reflects the deterministic propagation guard penalty of the T-timed Petri net formalism. This bound is independent of input contrast: regardless of the relative strength of $E_A$ and $E_B$, inhibitory suppression is guaranteed to arrive at the losing channel within 9 ms of stimulus onset. Input contrast determines the magnitude of suppression, not its timing. This constitutes a formal real-time guarantee on competition resolution, meaning the circuit cannot enter an indefinitely unresolved state. This property is not expressible in rate-coded or purely probabilistic neural models but follows directly from the schedulability analysis of the underlying Petri net.

	\subsection{Hierarchical Feature Detector}
\begin{figure}
		\centering
		\includegraphics{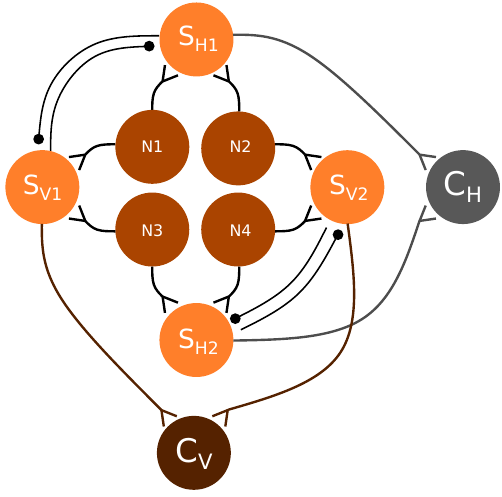}
		\caption{The hierarchical feature detector consisting of a $2\times 2$ grid, simple neurons $S$ that capture lines, and complex neurons $C$ that indicate the presence of horizontal and vertical lines.}
		\label{fig:HMAX}
	\end{figure}

	As a final example, we present a hierarchical feature detector for a $2\times 2$ grid, a toy example of the Hubel-Wiesel hierarchy formalized as the HMAX model for machine vision, shown in Fig. \ref{fig:HMAX}. The grid is composed of neurons N1--N4. Simple neurons S\textsubscript{H1} and S\textsubscript{H2} capture horizontal lines, while simple neurons S\textsubscript{V1} and S\textsubscript{V2} capture vertical lines. At the output of the network, complex neuron C\textsubscript{H} indicates the presence of horizontal lines while complex neuron C\textsubscript{V} indicates the presence of vertical lines. Simple neurons S\textsubscript{H$\alpha$} and S\textsubscript{V$\alpha$} mutually inhibit each other to reduce perseveration, the unintended continuation of detection of the previous pattern.
	
	The circuit exhibits graded orientation selectivity: preferred-orientation stimuli produce a 2:1 spike-count ratio between C\textsubscript{H} and C\textsubscript{V}, while diagonal (ambiguous) stimuli suppress both channels symmetrically. This soft winner-take-all behavior is consistent with the HMAX model and with biological orientation tuning curves. The selectivity index (SI) for constant proper neuronal excitations is 0.33.
	
	The WCRT for the first test (horizontal line pattern) was derived analytically by tracing the circuit's critical path. With all timing parameters set to unity and stimuly injected at every timestep, the first spike at the horizontal complex cell C\textsubscript{H} is predicted at $t=6$, decomposed as one timestep for accumulation across two input epochs, one synaptic delay from input to simple cell, one propagation period plus one timestep for the strict inequality guard at the simple cell, one synaptic delay from simple to complex cell, and one final accumulation step at C\textsubscript{H}. This yields the bound $WCRT=d_{N\rightarrow S}+(t_{prop}^N+1)+d_{S\rightarrow C}+(t_{prop}^S+1)=6$. Empirical measurement confirms the first peak at C\textsubscript{H} at $t=6$, with C\textsubscript{V} firing at $t=9$, a latency gap of $\Delta_\tau=3$ timesteps attributable entirely to inhibitory suppression from S\textsubscript{H1} onto S\textsubscript{V1} arriving at $t=4$. The strict inequality in the propagation guard introduces a deterministic one-timestep penalty per synaptic layer, a structural property of the T-timed Petri net formalism.
		
	\section{Conclusions}
	\label{sec:conclusions}
	
	In this work, we presented the Petri neuron, a formal model of the leaky integrate-and-fire neuron cast as a T-timed Petri net. We formalized the Petri neuron as a five-place, five-transition Petri net whose places encode accumulation, readiness, pre-spike propagation, output, and refractory recovery. The transitions implement input gating, threshold firing, axonal propagation, and leaky decay. Through structural analysis via spectral methods, we identified $t_{spike}$ as the dominant topological pivot and showed that the net is structurally live with no deadlocks. Moreover, using a Pad\'e approximation to the LIF state equation, we derived mapping rules that allow a designer to instantiate a Petri neuron from biological parameters directly. Using empirical jitter measurements, we showed that non-deterministic processors introduce timing errors, rendering them unsuitable for simulating neural circuits with sub-millisecond sampling periods.
	
	To demonstrate the model, we studied three microcircuits: feedback inhibition, lateral inhibition, and a hierarchical feature detector. The Petri neuron reproduces gamma-band rhythms and guarantees that inhibitory feedback arrives within a bounded number of cycles, preventing runaway excitation. The winning channel in the lateral inhibition circuit properly suppresses the loser within a formally bounded time window for competition resolution. The hierarchical feature detector, implemented on a $2\times 2$ grid with simple and complex cells, reproduces orientation selectivity with a selectivity index of 33\% and a 2:1 spike-count ratio between preferred and non-preferred orientations. The WCRT of the first response empirically matches the analytical prediction.
	
	Several limitations of the present work point toward directions for future research. First, the Petri neuron currently models only the absolute refractory period; the relative refractory period, during which the neuron remains excitable but requires a higher threshold, is not captured and would require and additional guarded place with a time-varying threshold. Second, the token-based accumulator discretizes the membrane potential into integer steps, introducing quantization error that grows at low firing rates and low synaptic weights; adaptive weight quantization of a multi-resolution accumulator could help mitigate this. Third, the current network model assumes static synaptic weights, whereas biological circuits rely heavily on short- and long-term synaptic plasticity; extending the formalism to include Hebbian or spike-timing-dependent plasticity (STDP) rules within the Petri net transition guards is a natural next step. Finally, the scalability of the event-driven scheduler to large networks has not been characterized; future work should establish complexity bounds on the priority queue and explore distributed execution strategies for deployment on multi-core or FPGA-based neuromorphic platforms.
		
	\bibliography{bib/main}

\end{document}